%% file: main.tex
\documentclass[runningheads]{llncs}

\usepackage[year=2024,ID=2529]{eccv}

\usepackage{eccvabbrv}

\usepackage{graphicx}
\usepackage{booktabs}
\usepackage{float}
\usepackage{algorithm}
\usepackage{algpseudocode}
\usepackage{eucal}

\usepackage[accsupp]{axessibility}  %

\def\K{\ensuremath{K}\xspace}
\def\V{\ensuremath{V}\xspace}

\usepackage{orcidlink}

\newcommand\blfootnote[1]{%
  \begingroup
  \renewcommand\thefootnote{}\footnote{#1}%
  \addtocounter{footnote}{-1}%
  \endgroup
}

\newcommand{\method}{FreeCompose\xspace}

\usepackage{hyperref}
\hypersetup{
    colorlinks=true,
    breaklinks=true,
    urlcolor=blue,
    linkcolor=blue,
    bookmarksopen=false,
    filecolor=black,
    citecolor=blue,
    linkbordercolor=blue
}

\begin{document}

\title{FreeCompose: Generic Zero-Shot \\Image Composition with Diffusion Prior} 

\author{
Zhekai Chen$^*$
\and
Wen Wang$^*$
\and
Zhen Yang
\and
Zeqing Yuan
\and
\\
Hao Chen
\and 
Chunhua Shen
}

\authorrunning{Accepted to European Conf. Comp. Vision (ECCV) 2024}

\institute{
Zhejiang University, China 
\blfootnote{
$^*$Equal contribution. HC and CS are the 
corresponding authors.
}
}

\maketitle

\input{sections/0.abs.tex}
\input{sections/1.intro.tex}

\input{sections/2.related_work.tex}

\input{sections/3.method.tex}

\input{sections/4.exp.tex}

\input{sections/5.conclusion.tex}
\input{sections/8.appendix.tex}

\bibliographystyle{alpha}
\bibliography{cvml,main2}
\end{document}

%% file: sections/0.abs.tex
\begin{abstract}

\begin{figure}[htp]
    \centering
    \vspace{-20pt}
    \includegraphics[width=0.985\textwidth]{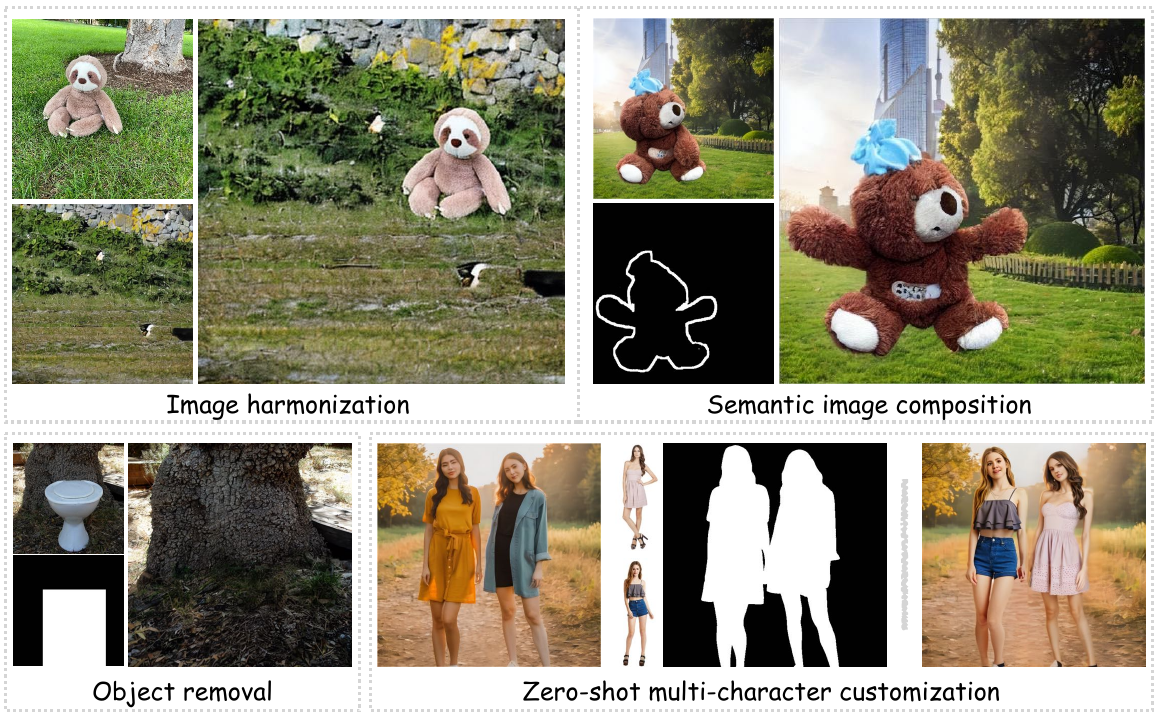}
    \caption{
    \textbf{\method}
    harnesses the generative prior of pre-trained diffusion models to achieve versatile image composition, such as appearance editing (image harmonization) and semantic editing (semantic image composition). Furthermore, it can be extended to various downstream applications, including object removal and multi-character customization.
    }
    \label{fig:teaser}
\vspace{-20pt}
\end{figure}

We offer a novel approach to image composition, which integrates multiple input images into a single, coherent image. Rather than concentrating on specific use cases such as appearance editing (image harmonization) or semantic editing (semantic image composition), we showcase the potential of utilizing the powerful generative prior inherent in large-scale pre-trained diffusion models to accomplish generic image composition applicable to both scenarios.
We observe 
that the pre-trained diffusion models automatically identify simple copy-paste boundary areas as low-density regions during denoising. 
Building on this insight, we propose to optimize the composed image towards high-density regions guided by the diffusion prior.
In addition, we introduce a novel mask-guided loss to further enable flexible semantic image composition.
Extensive experiments validate the superiority of our approach in achieving generic zero-shot image composition. 
Additionally, our approach shows promising potential in various tasks, such as object removal and multi-concept customization.

Project webpage: 
\url{https://github.com/aim-uofa/FreeCompose}

\keywords{Image composition \and Zero-shot \and Diffusion prior}
\end{abstract}

%% file: sections/1.intro.tex
\section{Introduction}\label{sec:intro}

Image Composition is a fundamental task in computer vision~\cite{tao2013error,zhu2015learning,tsai2017deep}, which aims to fuse the foreground object from one image with the background of another image to generate a smooth natural image. It has a wide range of applications in many fields, such as image restoration, art design, game development, virtual reality, and so on.

For this reason, a large amount of research has been conducted on image composition~\cite{tao2013error,zhu2015learning,tsai2017deep,dovenetharm}. Depending on whether there is a change in the semantic structure of the composite image, image composition can be broadly categorized as image harmonization~\cite{zhu2015learning,tsai2017deep} and semantic image composition~\cite{paintbyexample,chen2023anydoor}. The former modifies only the statistical information of the local area after pasting the foreground pixels into the background image, to obtain an image with a smooth transition between the front and background. In contrast, the latter fine-tunes the structure of the image according to the global image context and semantically blends the foreground and background.

As deep learning~\cite{lecun2015deep} gains its popularity, mainstream solutions for image composition adopt the learning-based pipeline~\cite{tsai2017deep,dovenetharm}. They require model training on data triplet of foreground, background, and composite images to achieve image combination. However, due to the difficulty in obtaining the triplets, these models can only be trained on a limited amount of training data with a specific data distribution, making it difficult to generalize to various scenarios in real-world applications.

In contrast, recent text-to-image diffusion models~\cite{ldm,DALLE2,imagen} have achieved large-scale pre-training using simple graphical data pairs, demonstrating strong generalization over open-world data distributions. Inspired by this, we attempt to utilize the image prior of the pre-trained diffusion model to realize generic image composition, in zero shot. Our key assumption is that the pre-trained diffusion model can accurately predict the noise component in natural images, while inaccurately for unnatural image regions that deviate from the pre-training data distribution. Based on this, we can localize the unnatural regions in a composite image after simply copying and pasting.

\begin{figure}[htp]
    \centering
    \includegraphics[width=\textwidth]{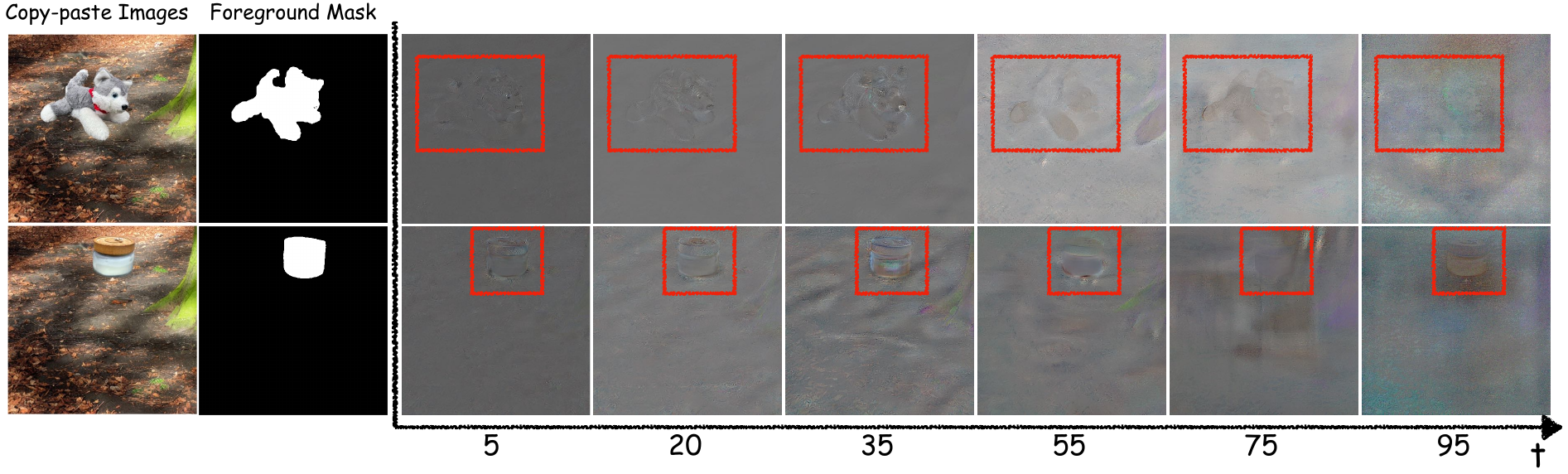}
    \caption{\textbf{
    Observations on the diffusion prior.
    } 
    The images on the left, denoted as copy-paste images, are obtained by simply pasting the foreground object to the background image.
    The frozen diffusion model takes the noisy copy-paste images from varying diffusion forward steps as input, and predicts the gradient to update the images (visualized on the right). 
    Low-density regions with larger gradient updates are highlighted by red boxes.
    The low-density regions are highly consistent with the inharmonious regions caused by %
    naive 
    copy-paste.
    }
    \label{fig:observation}
\vspace{-5pt}
\end{figure}

To validate this hypothesis, we conduct preliminary explorations on composite images, as shown in Figure ~\ref{fig:observation}.
Based on the above observations, we propose FreeCompose, which optimizes the pixels in the image such that it can be consistent with the image prior of the pre-trained diffusion model.

In our method, we aim to use the prior of the diffusion model to combine the object with the background without having to train the diffusion model itself (referred to as Training-free in this field). We propose a generic pipeline for composition that consists of three phases: object removal, image harmonization, and semantic image composition.
Unlike current works~\cite{duconet,paintbyexample} that rely on task-specific training for image harmonization or semantic image composition, our \method can directly utilize a pre-trained diffusion model and achieve composition in zero-shot.
During the object removal phase, our pipeline eliminates the foreground in the original image by manipulating the  \K, \V  values of the diffusion UNet's self-attention layer. In the image harmonization phase, the new object is combined with the background to create a harmonious scene.
If additional conditions for semantic image composition are provided, the composition is guided by the difference between the conditions, while preserving the object's identity through an additional 
replacement of the $K$, $V$ in the self-attention.

Based on these phases and techniques, \method can be effectively used for various tasks with promising results. These tasks include basic object removal, image harmonization, and semantic image composition. Moreover, \method demonstrates the ability to stylize objects by utilizing prompts during the image harmonization phase. Additionally, when combined with existing works, it can be applied to a wide range of tasks, such as multi-character customization.

To summarize, our contributions are listed as follows.
\begin{itemize}
    \item 
    Our findings indicate that the diffusion prior can automatically identify and focus on regions in the composite image that appear unnatural.
    
    \item Developing from the 
    vanilla 
    DDS loss, we explore and prove the possibility of additional designs for specific tasks including mask-guided loss and operations on  \K, \V  embeddings.
     These enhancements expand the range of applications for this loss format.
    \item FreeCompose achieves competitive results on both image harmonization and semantic image composition. Moreover, it facilitates broad applications including object removal and multi-character customization.
    \item 

    In contrast to existing methods that train separate models for individual image composition problems, the diffusion prior that we use offers a generalized natural image prior that can effectively perform both image harmonization and semantic image composition in a zero-shot manner.
    
\end{itemize}

%% file: sections/2.related_work.tex
\section{Related Work}\label{sec:related}

\textbf{Image Harmonization}
Image harmonization aims to generate a realistic combination of foreground and background contents from different images. 
It focuses on adjusting low-level appearances, like the global and local color distribution change caused by light and shadows, while maintaining the content structure unchanged.
Early works on image harmonization\cite{pitie2005n, cohen2006color, reinhard2001color, tao2013error, sunkavalli2010multi} rely on hand-crafted priors on color~\cite{pitie2005n}, gradient~\cite{tao2013error}, or both~\cite{sunkavalli2010multi}.
With the advance of deep learning~\cite{lecun2015deep}, recent methods~\cite{zhu2015learning,tsai2017deep,dovenetharm,jiang2021ssh,Ling2021region,Hao2020ImageHW,cong2020bargainnet,Sofiiuk2021fore,chen2024deep} explore learning-based methods for image harmonization.
For example, Zhu \etal \cite{zhu2015learning} train a discriminative model to judge the realism of a composited image, and leverage the model to guide the appearance adjustment of a composed image.
Tsai \etal ~\cite{tsai2017deep} propose the first end-to-end network for image composition.
Subsequently, DoveNet~\cite{dovenetharm} leverages a domain verification discriminator to migrate the domain gap between the foreground and background images.
Recently, Tan \etal \cite{duconet} proposed a new end-to-end net named DocuNet by leveraging the channels of images and achieved excellent success. 
While effective, these image harmonization models are trained on domain-specific datasets, and struggle to generalize to open-world images.
By contrast, we leverage the natural image prior preserved in large-scale pre-trained diffusion models for zero-shot image harmonization in the wild.
Chen \etal  \cite{huang2024diffusion} also attempted to use diffusion model as a base model for harmonization by a method called Diff-harmonization composed of inversion and re-denoising, but limited to harmonization.

\textbf{Image Editing} Text editing is a broad area that encompasses many research topics, including image-to-image translation~\cite{isola2017image, zhu2017unpaired, kim2017learning, mustafa2020transformation, zhang2018meta}, inpainting~\cite{lugmayr2022repaint, multiconcept, lizuka2017globally, pathak2016context, yang2016highresolution, gao2017on, liu2018image}, text-driven editing~\cite{hertz2022prompt, xia2020tedigan, blendeddiffusion, patashnik2021styleclip, tao2022denet}, etc.
We refer readers to \cite{huang2024diffusion,zhan2023multimodal} for more comprehensive review. 
Here we focus on the image inpainting task.
Traditional image inpainting takes the masked image as input, and predicts the masked pixels from the image context. 
For example, LaMa~\cite{suvorov2022resolution} enlarges the receptive fields from the perspective of both modeling and losses, thus achieving inpainting in large masks and complex scenarios.
Recently, benefiting from large-scale pre-trained text-to-image generative models~\cite{DALLE2,ldm}, researchers explore additional text input to guide the inpainting process~\cite{avrahami2022blended,avrahami2023blended}.  
For example, Blended Latent Diffusion~\cite{avrahami2023blended} proposes to smoothly blend the latent of the foreground region and the background areas to achieve text-guided inpainting.
Another line of work~\cite{paintbyexample,chen2023anydoor} inpaints the masked image with an example image, which is also known as semantic image composition~\cite{paintbyexample}. 
Different from image harmonization which only alters low-level statistics, semantic image composition semantically transfers the foreground object (often with structural changes) during composition.
A representative work Paint-by-Example~\cite{paintbyexample} fine-tunes the pre-trained Stable Diffusion model to take additional exemplar images as input for inpainting.
AnyDoor~\cite{chen2023anydoor} improves the semantic image composition pipeline to preserve the texture details in exemplar images and leverage the multi-view information in video datasets for effective training.

{\bf Diffusion Models}
Diffusion models~\cite{ho2020denoising,sohl2015deep} have emerged as powerful generative models for images.
Large-scale pre-trained diffusion models, like DALLE-2~\cite{DALLE2}, Imagen~\cite{imagen}, Stable Diffusion~\cite{ldm}, and SDXL~\cite{podell2023sdxl}, demonstrate unprecedented text-to-image generation capacities in terms of both realism and diversity.
Motivated by the success of diffusion models, attempts have been made to leverage pre-trained image diffusion models as the prior for other generative tasks~\cite{poole2022dreamfusion,wang2024prolificdreamer,katzir2023noise}.
Considering the data scarcity of 3D assets, DreamFusion~\cite{poole2022dreamfusion} uses Imagen~\cite{imagen} as a generative prior, and proposes a novel Score Distillation Sampling (SDS) loss for optimizing the implicit representation of a 3D object. 
Subsequently, ProlificDreamer~\cite{wang2024prolificdreamer} models the parameters of 3D assets as a random variable and proposes the variational score distillation to alleviate the over-saturation and over-smoothness in DreamFusion.
Different from these works that focus on text-to-3D generation, DDS~\cite{hertz2023delta} tackles the task of text-guided image editing, and identifies the editing region by referencing the original image and its corresponding prompt.
In this work, we also leverage diffusion models as the generative prior (diffusion prior). 
Our key observation is that diffusion prior helps locate unnatural areas in simple copy-paste image composition. 
Based on this, a masked guided loss is proposed to enable generic smooth image composition.

%% file: sections/3.method.tex
\section{Preliminaries}\label{sec:preliminaries}

\subsection{DDS Loss}

The Delta Denoising Score (DDS) \cite{hertz2023delta} is developed froma modification of the diffusion loss and Score Distillation Sampling \cite{dreambooth} for image editing. Given an input image $I$, the diffusion model encodes it into a latent variable $\mathbf{z}$. Using a prompt $P$ for theto generation ofe a text embedding $y$, a timestep $t$ is randomly chosen from a uniform distribution $\mathcal{U}(0,1)$, and noise $\epsilon$ is sampled from a normal distribution $\mathcal{N}(0, \mathbf{I})$. A noised latent variable $\mathbf{z_t}$ can then be represented as $\mathbf{z_t}=\sqrt{\alpha_t}\mathbf{z}+\sqrt{1-\alpha_t}\epsilon$, where $\alpha_t$ is determined by a noise scheduler based on $t$.

Given a pre-trained diffusion model $\epsilon_\phi$ with parameter set $\phi$, a modified predicted noise according to classifier-free guidance \cite{ho2022classifierfree} can be expressed as
$$\epsilon^w_\phi(\mathbf{z_t}, y, t)=(1+w)\epsilon_\phi(\mathbf{z_t},y,t)-w\epsilon_\phi(\mathbf{z_t},t),$$
where $\epsilon_\phi(\mathbf{z_t},y,t)$ is the raw noise predicted by the diffusion model conditioned on $y$, $\epsilon_\phi(\mathbf{z_t},t)$ is unconditioned noise, and $w$ is a weight for balance.

Using two image-text pairs $I_i, P_o$ and $I_t, P_t$, the DDS loss with respect to parameter $\theta$ can be expressed in gradient form as:
\begin{equation}\label{eq:01}
\nabla_\theta\mathcal{L}_{\rm DDS}=(\epsilon^w_\phi(\mathbf{z_t}, y, t)-\epsilon^w_\phi(\mathbf{\hat{z_t}}, \hat{y}, t))\frac{\partial\mathbf{z_t}}{\partial\theta},
\end{equation}
where $\epsilon^w_\phi(\mathbf{z_t}, y, t)$ is predicted from $I_i, P_o$ and $\epsilon^w_\phi(\mathbf{\hat{z_t}}, \hat{y}, t)$ is predicted from $I_t, P_t$ with the same $t$ and $\epsilon$. For simplicity, this loss is denoted as $\mathcal{L}_{\rm DDS}(I_i, I_t, P_o, P_t)$.

\subsection{Perceptual Loss}

The perceptual loss \cite{johnson2016perceptual} is proposed to measure the perceptual similarity of images based on the features of VGG-16 \cite{simonyan2014deep}. 
Although originally designed for the super-resolution task by maintaining the features of the original image, it also allows for the preservation of selected regions. 
We denote the perceptual loss between $I_i$ and $I_t$ as $\mathcal{L}_{\rm per}(I_i, I_t)$.

\section{Method}\label{sec:method}

Given a target image $I_t$ with the object's mask $M_t$ and a background image $I_s$ with a designated region $M_s$ for placing the object, our goal is to compose a new coherent image that retains the background from $I_s$ while incorporating the target image's object as the foreground.

To achieve generic image composition, our method comprises three phases: \textbf{object removal}, \textbf{image harmonization}, and \textbf{semantic image composition}.
This design allows for the composition of various foreground object and background images.
In Figure  \ref{fig:method_overview}, we illustrate the pipeline with special segments of different phases. The overview of the pipeline is presented in \cref{overall_pipeline}, followed by details of object removal in \cref{subsec:object_removal}, image harmonization in \cref{subsec:image_harmonization}, and semantic image composition in  \cref{subsec:semantic_image_composition}.

\subsection{Overall pipeline}\label{overall_pipeline}

The removal stage takes $I_s$ and $M_s$ as inputs to generate a background image $I_b$ with the object in $M_s$ removed. Subsequently, the composition stage produces a coherent image $I_c$ given $I_b, M_s, I_t, M_t$. Furthermore, if conditions are provided to transfer the object from the original condition $C_o$ to the target condition $C_t$, the editing stage can integrate these conditions onto $I_c$ to synthesize image $I_{res}$. The conditions can take the form of text or other formats accepted by T2I-Adapter\cite{mou2023t2iadapter}. Each stage is optimized with different loss functions: $\mathcal{L}_{\rm rmv}, \mathcal{L}_{\rm ham}$, and $\mathcal{L}_{\rm com}$.

\begin{figure}[t]

\centering

\includegraphics[width=1.0\textwidth]{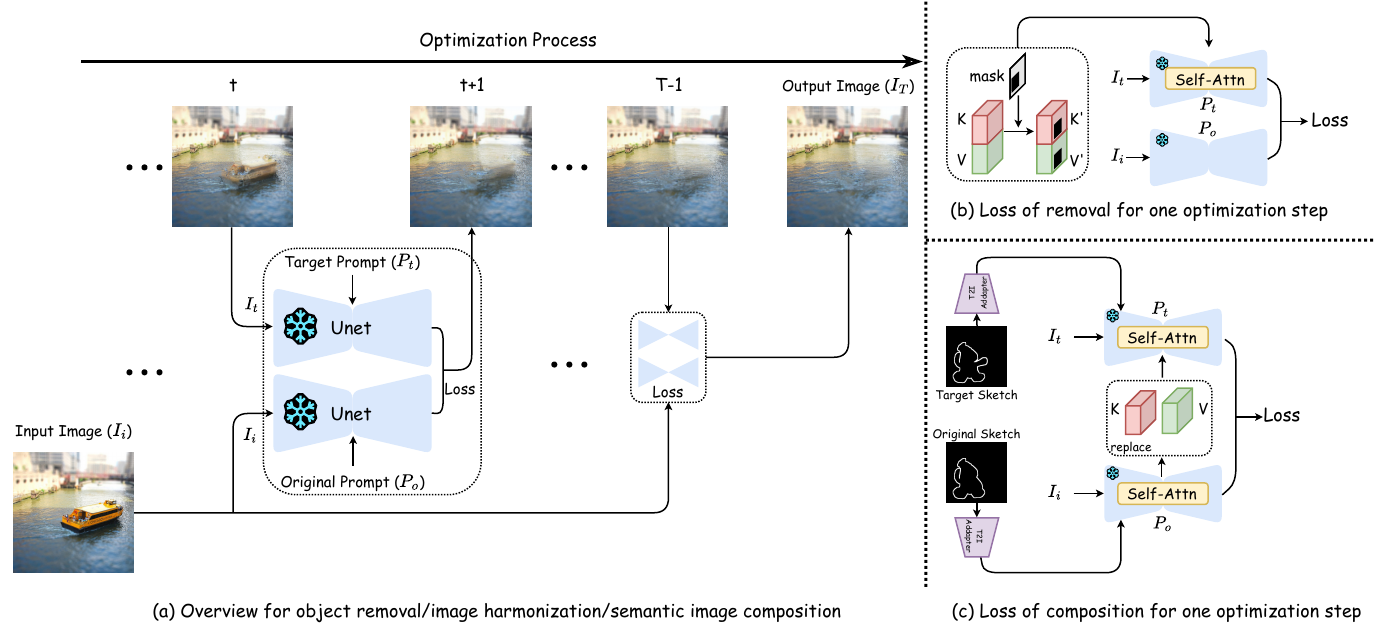}

\caption{\textbf{Pipeline overview.} 
Our 
FreeCompose pipeline consists of three phases: object removal, image harmonization, and semantic image composition. In each phase, the pipeline takes an input image and two text prompts to calculate the loss. In the object removal phase, an additional mask is required to select  \K, \V  values. In the semantic image composition phase, text prompts can be replaced by other formats, and an additional  \K, \V  replacement is implemented for identity consistency.}
\label{fig:method_overview}

\end{figure}

The method follows a general pipeline across all three phases, as depicted in Figure  \ref{fig:method_overview}. With inputs including an image $I_i$, an original prompt $P_o$, and a target prompt $P_t$, the pipeline initializes with an optimized image $I_t$ and guides its progression to the output image $I_T$ using a phase-specific loss function.

$P_o$ and $P_t$ are set as general prompts for object removal and image harmonization as elaborated in \cref{subsec:object_removal} and \cref{subsec:image_harmonization}, whereas for semantic image composition, they are taken as input conditions.

In general, the DDS loss can modify images but may also distort them during optimization. Meanwhile, the perceptual loss helps maintain object identity. When used together, these losses can create a balanced loss function that forms the backbone of the pipeline. At the same time, minimum adjustment to the loss function enables other specific tasks, as detailed in the following sections.

\subsection{Object Removal}\label{subsec:object_removal}

In this phase, we take $I_s$ as the input image $I_i$, and the object region mask $M_s$ is required. $P_o$ and $P_t$ are set as placeholder prompts, such as ``Something in some place'' and ``Some place,'' when no prompt is provided. These prompts are partially effective, but they do not have the capability to directly eliminate the object, as shown in the ablation study (see  \cref{fig:removal_ablation}).

We add an extra segment during the calculation of the DDS loss to enhance the ability of removal, as shown in Figure  3(b). 
The diffusion model is based on a UNet architecture, composed of residual, self-attention and cross-attention blocks. In the self-attention blocks, features are projected into quires $Q$, keys $K$ and values $V$, and the output can be represented as:
\begin{equation}
    \text{Attention}(Q, K, V) = \text{Softmax}\bigl( \frac{QK^T}{\sqrt{d}} \bigr) V,
\end{equation}
where $d$ is the dimension of the hidden states.

Based on previous work such as \cite{huang2023KV}, the  \K, \V  values of the self-attention layer during the denoising step are observed to have an effect on the semantic result. Guided by this discovery, we use $M_s$ to discard some  \K, \V  values partially. Specifically, for a $K$ or $V$ value of shape $B\times l\times d$, where $B,l,d$ represent the batch size, sequence length, and input dimension, respectively, we resize the mask to shape $h\times w = l$ and flatten it to a sequence with length $l$. By selecting indices from $v_i > \text{threshold}$, where $v_i$ represents the value of index $i$ in the sequence, the semantic information of the masked region is replaced by its surroundings, thereby achieving the objective of removal. This mask guided loss can be represented as $\mathcal{L}^{\rm rmv}_{\rm DDS}(I_i, I_t, P_o, P_t, M)$ with the gradient form:

$$\nabla_\theta\mathcal{L}^{\rm rmv}_{\rm DDS}=(\epsilon^w_\phi(\mathbf{z_t}, y, t)-\epsilon^w_\phi(\mathbf{\hat{z_t}}, \hat{y}, t, M))\frac{\partial\mathbf{z_t}}{\partial\theta},$$

The only difference with Eq. \ref{eq:01} is the $\epsilon^w_\phi(\mathbf{\hat{z_t}}, \hat{y}, t, M)$, which means that the  \K, \V  values of the self-attention layers masked by $M$ are excluded during noise prediction.

The %
overall 
loss function, thus, comprises two terms:
\begin{align}\label{eq:02}
\mathcal{L}_{\rm rmv}=\mathcal{L}^{\rm rmv}_{\rm DDS}(I_s, I_t, P_o, P_t, M_s)+\lambda_{\rm per}\mathcal{L}_{\rm per}(I_s \otimes M_s', I_t\otimes M_s').
\end{align}
Here, $M_s'$ is the reversed mask of $M_s$, $\otimes$ denotes the Hadamard production of two images, and $\lambda_{\rm per}$ is a hyperparameter used to balance the two losses.

\subsection{Image Harmonization}\label{subsec:image_harmonization}

Applying the bounding box of $M_s$, a copy-paste image $I_p$ and its corresponding object mask $M_p$ can be obtained from $I_b, I_t, M_t$. This image is used as input image $I_i$ ($I_i=I_p$) in this phase. Without designated prompts, an empty prompt and ``A harmonious scene'' are initialized as $P_o, P_t$ for the DDS loss. The perceptual loss is used separately for the background and the foreground to preserve background appearance and object identity. The overall loss consists of three 
terms: 
\begin{equation}
\label{eq:03}
\begin{array}{ll}
    \mathcal{L}_{\rm har}=&\mathcal{L}_{\rm DDS}(I_p, I_t, P_o, P_t)\\
    &+\lambda_{\rm bak}\mathcal{L}_{\rm per}(I_p\otimes M_p', I_t\otimes M_p')+\lambda_{\rm for}\mathcal{L}_{\rm per}(I_p\otimes M_p, I_t\otimes M_p),
\end{array}
\end{equation}
where $M_p'$ is the revered mask of $M_p$, $\lambda_{\rm bak}$ is a hyperparameter used to balance the perceptual loss related to the background and  $\lambda_{\rm for}$ is a hyperparameter used to balance the perceptual loss related to the target object.

\subsection{Semantic Image Composition}\label{subsec:semantic_image_composition}

This phase accepts either the copy-paste image $I_p$ or the composition result $I_c$ and requires two additional conditions: $C_o$ and $C_t$. If the conditions are in text form, they will be directly used as $P_o$ and $P_t$ for the DDS loss. Conditions in other forms will be translated by T2I-Adapter\cite{mou2023t2iadapter} and added to the diffusion UNet as shown in Figure  \ref{fig:method_overview}(c).

An additional design is employed to maintain the identity of the object during the editing procedure. As displayed in Figure  \ref{fig:method_overview}(c), \method replaces the optimized image $I_t$'s  \K, \V  values with $I_i$'s  \K, \V  values during the calculation of DDS loss. Specifically, for a DDS loss with $K_i, V_i, K_t, V_t$, where $K_i, V_i$ represent the  \K, \V  values of $I_i$, and $K_t, V_t$ represent the  \K, \V  values of $I_t$, we modify the calculation of self-attention in the diffusion UNet concerning $I_t$ as follows:
$$\left\{
    \begin{array}{ll}
        \text{Attention}(Q, K_i, V_i), &\text{if} t>T \text{and} l>L \\
        \text{Attention}(Q, K_t, V_t), &\text{otherwise,}
    \end{array}
\right.$$
where $t$ is the count of optimization, $l$ is the layer index of the self-attention layer, $T$ and $L$ are hyperparameters indicating the count number and layer index of self-attention to start such replacement. Because the background is also preserved along with the replacement, no perceptual loss is required. Therefore, the complete loss has the same format as the DDS loss:
$$\mathcal{L}_{\rm com} = \mathcal{L}^{\rm com}_{\rm DDS}(I_c, I_t, C_o, C_t),$$
where $\mathcal{L}^{\rm com}_{\rm DDS}(I_c, I_t, C_o, C_t)$ represents the DDS loss using $C_o, C_t$ as substitutes for conditions in forms besides text, with an additional design of  \K, \V  replacement during calculation.

%% file: sections/4.exp.tex
\section{Experiments}\label{sec:exp}

\begin{table}[t]
\centering\footnotesize
\setlength{\tabcolsep}{8pt}
\caption{%
\textbf{Results of User Study on Object Removal.}
The participants are requested to evaluate the results based on two aspects: (1) the level of image harmony after the object has been removed, and (2) the extent to which the object has been completely removed. Each criterion is rated from 1 (worst) to 5 (best) without additional explanation.
}
\label{tab:removal_user_study}
\begin{tabular}{lcc}
\toprule
& Image Harmony $\uparrow$& Object Removal $\uparrow$\\
\midrule
Repaint\cite{lugmayr2022repaint}& $3.24\pm 1.23$& $3.82\pm 1.35$\\
SD Inpainting\cite{multiconcept}& $2.99\pm 1.37$& $3.55\pm 1.34$\\
Lama\cite{suvorov2022resolution}& $3.47\pm 1.16$& $4.14\pm 0.94$\\
\method (ours)& $\mathbf{3.85 \pm 1.01}$&  $\mathbf{4.47\pm 0.73}$\\
\bottomrule
\end{tabular}
\end{table}

\begin{table}[b]
\caption{%
\textbf{Results of User Study on Image Harmonization.}
Participants are asked to rate the results based on (1) image harmony after the composition of the object, and (2) how well the identity of the object is preserved. 
}
\label{tab:composition_user_study}
\centering\footnotesize
\setlength{\tabcolsep}{8pt}
\begin{tabular}{lcc}
\toprule
& Image Harmony$\uparrow$& Object Identity Preserving$\uparrow$\\
\midrule
Diff Harmonization\cite{huang2024diffusion}& $3.11 \pm 1.04$& $3.83 \pm 1.10$\\
DucoNet\cite{duconet}& $3.14 \pm 1.17$& $\mathbf{4.16 \pm 1.04}$\\
\method (ours)& $\mathbf{3.69 \pm 1.07}$& $4.11 \pm 0.92$\\
\bottomrule
\end{tabular}
\end{table}

\subsection{Implementation Details}

\subsubsection{Global Hyperparameters.}
We use Stable Diffusion V2.1\footnote{https://huggingface.co/stabilityai/stable-diffusion-2-1} as the pre-trained model for real images, and AnyLoRA\footnote{https://huggingface.co/Lykon/AnyLoRA} as the pre-trained model for anime and cartoon images. We align the resolution of input images with the diffusion model to $512\times 512$. The Adam optimizer is adopted with a fixed learning rate of $5e^{-2}$.

\subsubsection{%
Hyperparameters.}
In the object removal phase, the DDS loss outside the mask resized from $M_s$ to the latent size is multiplied by 0.2 to limit the transformation of the background. Additionally, $\lambda_{\rm per} = 0.3$. In the image harmonization phase, $\lambda_{\rm bak}=0.3$ and $\lambda_{\rm for}=0.1$. The semantic image composition only uses the DDS loss with $T=400$ and $L=10$ for the replacement design.

\subsubsection{Prompt Usage.}\label{subsec:prompt_usage}
Two prompts, $P_o$ and $P_t$, are required for every calculation of the DDS loss. Providing specific prompts will improve the optimization procedure.
Our FreeCompose does not rely on user-provided text prompts for image composition. Instead, we predefined the prompts for different phases.
Specifically, in the object removal phase, we set $P_o$ as ``Something in some place.'' and  $P_t$ as ``Some place.'', respectively.
Similarly, we adopt empty prompts 
for $P_o$ and ``A harmonious scene.'' for $P_t$ in the image harmonization phase.
These prompts have proven to be effective.

\subsection{Main Results}

\subsubsection{Object removal}
\begin{figure}[htp]
\centering
\includegraphics[width=0.8\textwidth]{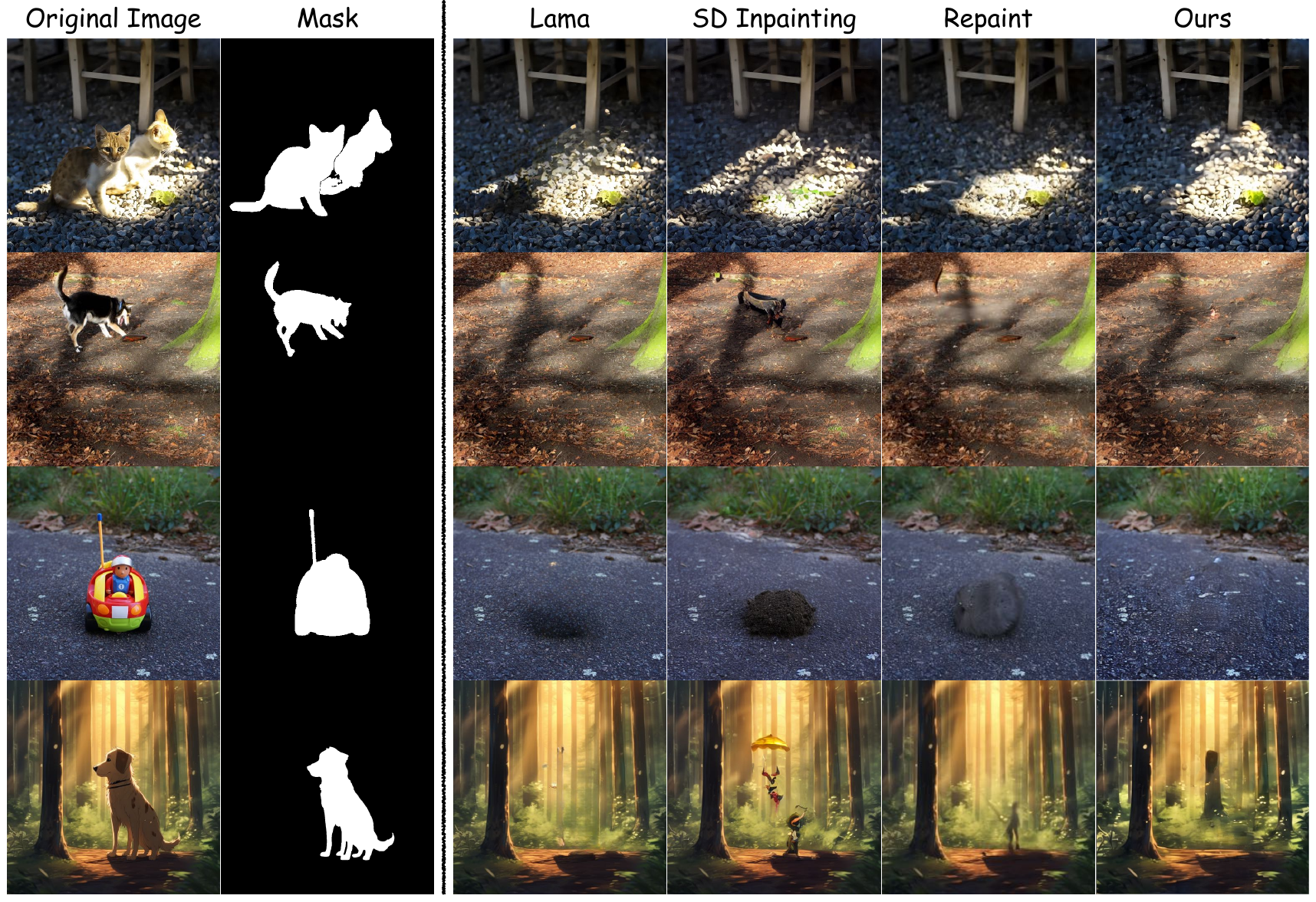}
\caption{\textbf{Qualitative comparison on object removal.} We compare with Lama\cite{suvorov2022resolution}, Stable Diffusion Inpainting\cite{multiconcept}, and Repaint\cite{lugmayr2022repaint}.
}
\label{fig:removal_comparison}
\end{figure}
In \cref{fig:removal_comparison}, we present the results of object removal, comparing them with previous work on removal and inpainting. When using the default prompts in \cref{subsec:prompt_usage}, Lama \cite{suvorov2022resolution}, Stable Diffusion Inpainting \cite{multiconcept}, and Repaint \cite{lugmayr2022repaint} require the same input as our method. This includes one original image along with a corresponding mask for the region that needs to be removed. As shown, SD Inpainting and Repaint struggle to completely remove the object, leaving some parts unchanged or replaced by something that doesn't fit well, like the outline of the dog in the second case and the unknowns in the fourth case. Although Lama performs better in removing the object and reconstructing the background, it fails to remove certain attachments of the object, such as the shadow in the third case. In general, our method demonstrates a stronger capability in removing the object and seamlessly filling the resulting areas, as can be observed in the third case where other methods perform poorly.

\subsubsection{Image harmonization}
\begin{figure}[t!]
    \centering
    \includegraphics[width=0.8\textwidth]{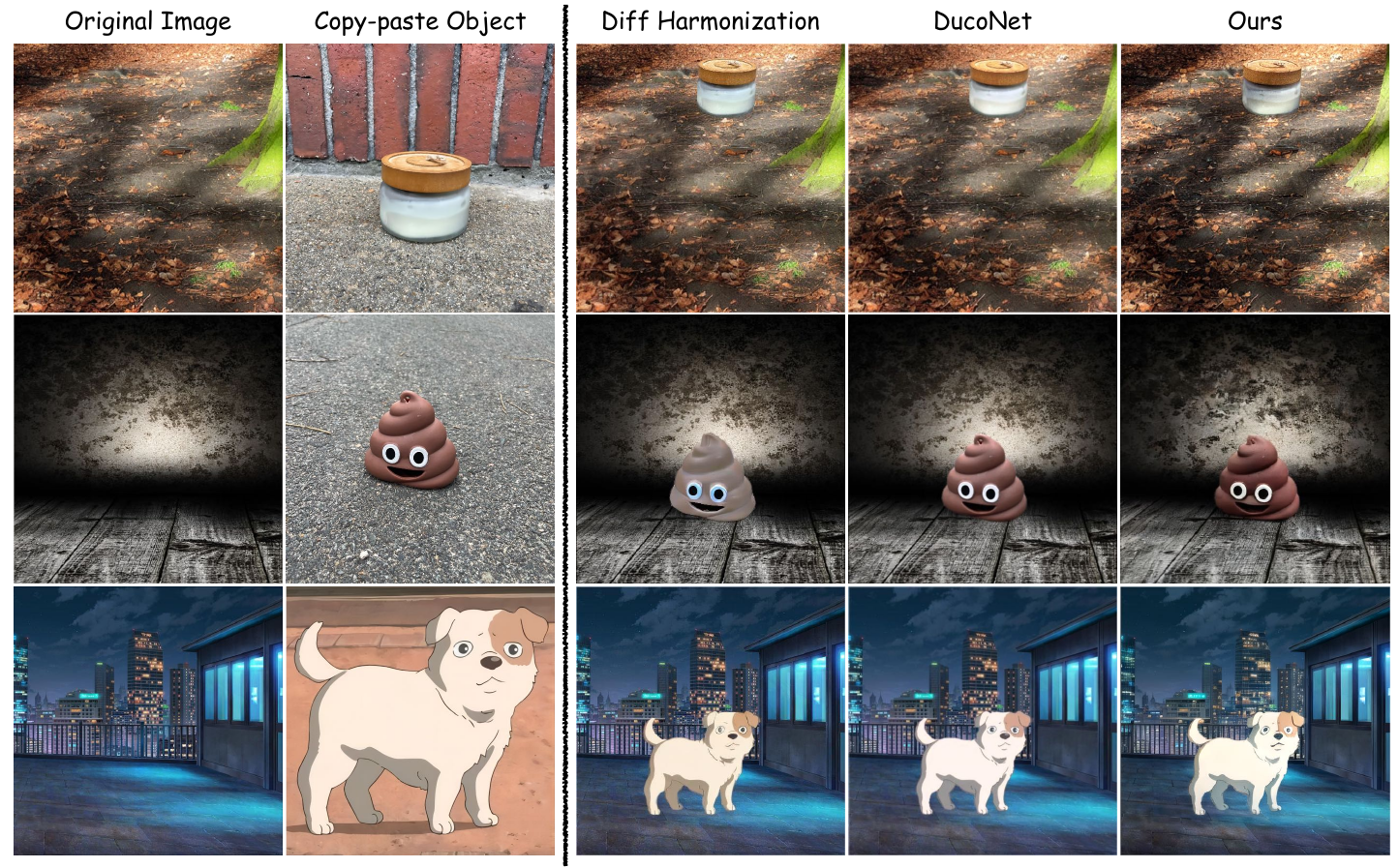}
    \caption{\textbf{Qualitative comparison on image harmonization.} We compare our method with zero-shot Diff Harmonization \cite{huang2024diffusion} and training-based DucoNet \cite{duconet}. 
    }
    \label{fig:composition_comparison}
\end{figure}
As shown in Figure  \ref{fig:composition_comparison}, Diff Harmonization successfully generates primary shadow as surface variation in the first candle case. However, it struggles to retain identity features such as the color of the second emoji case and the shape of the third dog case's eye. On the other hand, DucoNet preserves these features well but lacks realistic shadow and light effect under certain environments. For instance, in the first case, DucoNet simply illuminates the entire object, without accurately transforming the dark and bright sections according to the original image. In contrast, our method is capable of both preserving the object's identity and generating realistic lighting effects. For example, in the first case, \method enables the object to be covered by the shadow of the surroundings, resulting in the corresponding dark section while maintaining the object's identity.

\subsubsection{Semantic image composition}
\begin{figure}[t]
\centering
\includegraphics[width=1\textwidth]{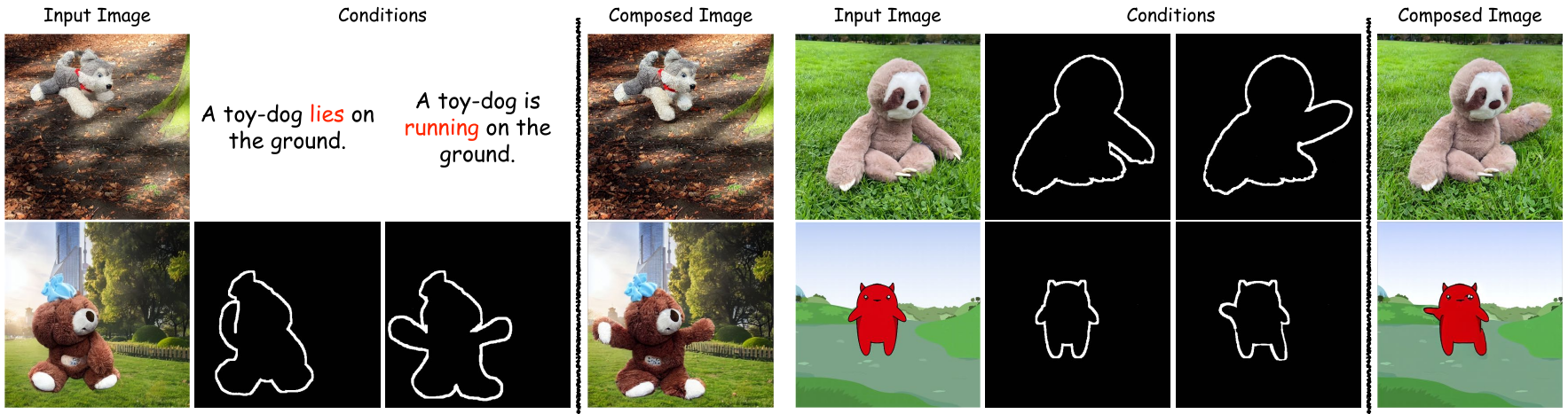}
\caption{\textbf{
Results on semantic image composition.} Our method accepts various conditions as guidance, including text and sketches. The case in the top-left corner uses different prompts as guidance for editing, while the other cases are guided by different sketches with identical prompts.}
\label{fig:editing_result}
\end{figure}
Figure ~\ref{fig:editing_result} illustrates the results of our semantic image composition. By using an input image (either a copy-paste image or an image after harmonization), \method is able to generate a composed image that maintains semantic consistency, guided by the disparity between two input conditions. As shown, the top-left case makes use of the difference between two prompts to transfer the dog from a lying posture to a running posture. In other cases, with the same prompt during calculation, the features extracted by T2i-Adapter \cite{mou2023t2iadapter} from different sketch images serve as guidance for semantic composition, proving the feasibility of wider usage.

\subsubsection{Quantitative comparison.}
Since our primary focus is on open domain questions, we believe that evaluating performance through user studies is more appropriate. We have planned a user study to assess the results of object removal and image harmonization, comparing them with previous works with five cases respectively. The study involves more than twenty volunteers.
To evaluate the effectiveness of object removal, participants are asked to assess the outcomes based on two criteria: (1) the level of image harmony achieved after the object is removed, and (2) the extent to which the object removal is executed. In terms of image harmonization, participants are instructed to assess the results based on two aspects: (1) the level of visual coherence achieved after integrating the object into the composition, and (2) the degree to which the object's identity is preserved. Each metric is rated on a scale ranging from 1 to 5.

The results are shown in Table  \ref{tab:removal_user_study} and Table  \ref{tab:composition_user_study}. As demonstrated, our method excels in both aspects for object removal. Our method in image harmonization received the highest rating for ``Image Harmony'', but it lagged behind DucoNet in terms of ``Object Identity Preservation.'' One possible reason is that our method employs a stronger composition strategy by restricting the weight of the foreground loss, resulting in partial degradation of the object's identity.

\subsection{Ablation Study}

\begin{figure}[t]
\centering
\includegraphics[width=0.98\textwidth]{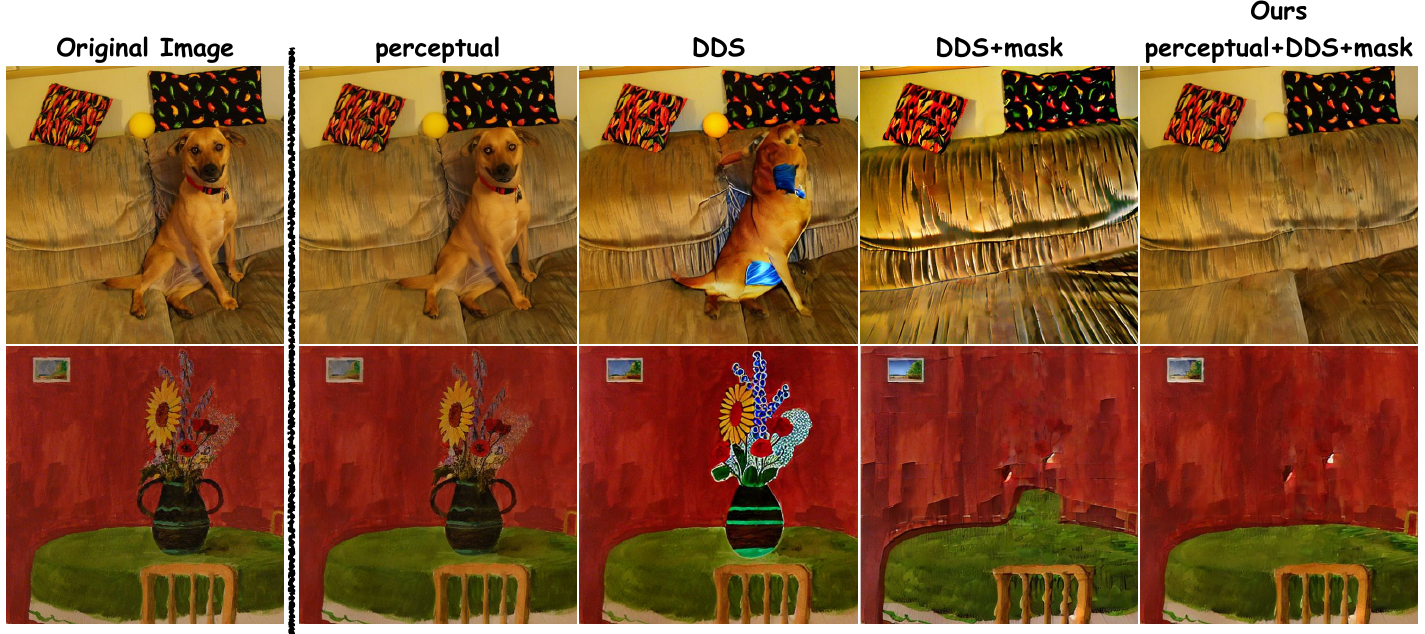}
\caption{\textbf{Qualitative ablation studies on} loss sections of the object removal phase. ``perceptual'' refers to the perceptual loss section in Eq.~\eqref{eq:02} alone, ``DDS'' refers to the %
vanilla 
DDS loss, and ``DDS+mask'' refers to the mask-guided DDS loss in  \cref{subsec:object_removal}.}
\label{fig:removal_ablation}
\end{figure}

We conducted an ablation study to validate our designs and analyze their functions. by disassembling and visualizing each design to clearly demonstrate their effects.

\subsubsection{Object Removal Phase.}

In Figure  \ref{fig:removal_ablation}, designs of the object removal phase are disassembled for analysis. The perceptual loss alone maintains the original image without any changes as the the ``perceptual'' column displayed. When using a raw DDS loss with default prompts, the object cannot be completely eliminated, resulting in some variations in the object in line with the ``DDS'' column. The introduced mask design in \S \ref{subsec:object_removal}, which selectively discards specific KV values based on the mask of the object, overcomes this limitation and enables the loss to successfully remove the object. However, such mask guided loss affects the background which should be preserved, as presented in the ``DDS+mask'' column. The last addition of perceptual loss section helps preserve the background while calculating the mask guided loss and generates the background image independently from the original foreground as demonstrated in the ``Ours'' column.

\subsubsection{Image Harmonization Phase.}

\begin{figure}[htp]
\centering
\includegraphics[width=0.988\textwidth]{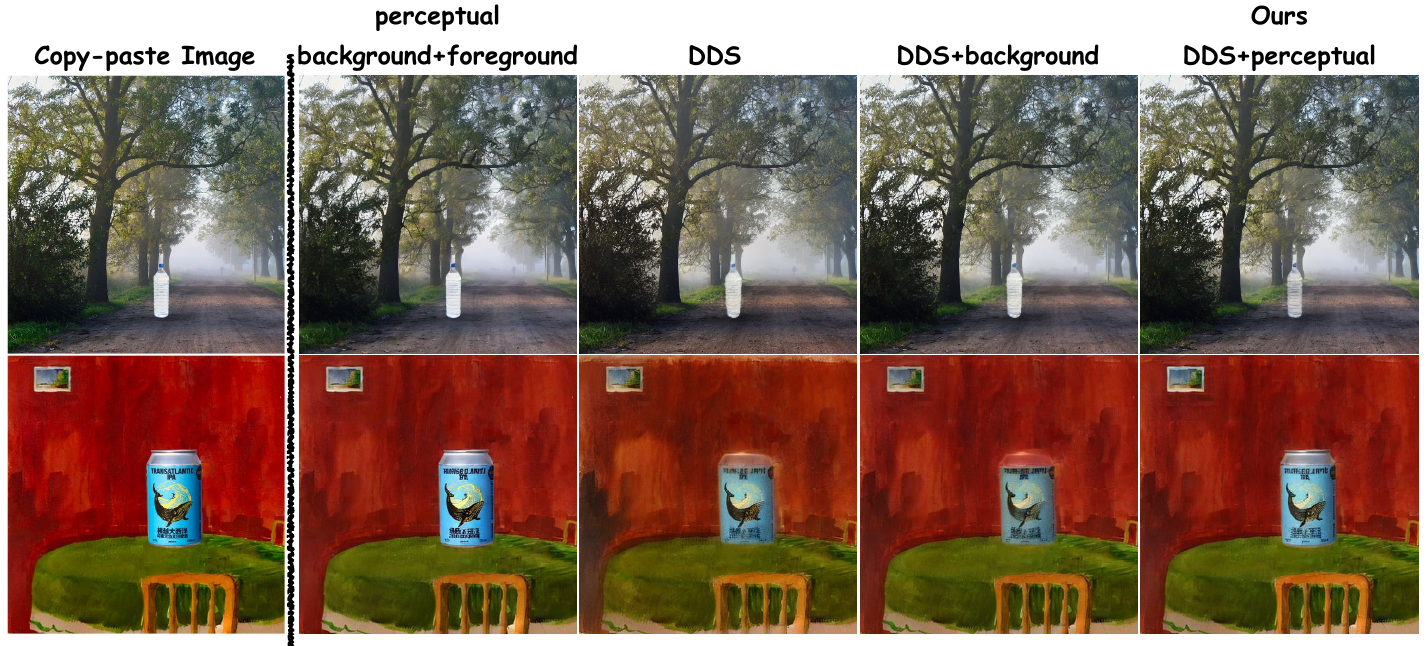}
\caption{\textbf{Qualitative ablation studies on} loss sections of the image harmonization phase. ``background'' refers to the background perceptual loss in Eq. \eqref{eq:03}, ``foreground'' refers to the foreground perceptual loss in Eq. \eqref{eq:03}, ``perceptual'' represents the sum of ``background'' and ``foreground'', and ``DDS'' represents the DDS loss.}
\label{fig:composition_ablation}
\end{figure}
In Figure  \ref{fig:composition_ablation}, different sections of the loss are ablated for observation of their respective functions. The perceptual loss, comprising the background perceptual loss and the foreground perceptual loss, ensures the consistency with the original copy-paste image, as seen in the ``perceptual'' column. When using the raw DDS loss, it allows for seamless blending of the object with the background, but may unintentionally remove certain features from both the foreground and the background, compatible with the ``DDS'' column. By employing distinct perceptual loss functions for the foreground and the background, the trade-off among the degree of integration, the identity of the object and the features of the background is achieved, enabling the generation of a harmonious image as shown in the ``Ours'' column.

%% file: sections/5.conclusion.tex
\section{Conclusion}
\label{sec:conclusion}

We present \method, a generic zero-shot image composition method that utilizes diffusion prior. In this work, we noticed that pre-trained diffusion models are capable of detecting inharmonious portions in copy-paste images. Building on this observation, we successfully apply this prior to both image harmonization and semantic image composition. \method is a zero-shot method, allowing easy usage without additional training. Moreover, it's suitable for various applications, showcasing the potential of the diffusion model prior.

We believe that the prospect of diffusion prior extends beyond what we have achieved thus far. In the future, we plan to explore additional uses for other composition tasks and to apply our method to video, capitalizing on its full capabilities.

\section*{Acknowledgement}

This work was supported by National Key R\&D Program of China (No.\  20\-2\-2\-Z\-D\-0\-1\-18\-700). The
authors would like to thanks Hangzhou City University for accessing its GPU cluster.

%% file: sections/8.appendix.tex
\appendix

\section*{Appendix}

\section{More Applications}

\begin{figure}[htp]
\centering
\includegraphics[width=1.0\textwidth]{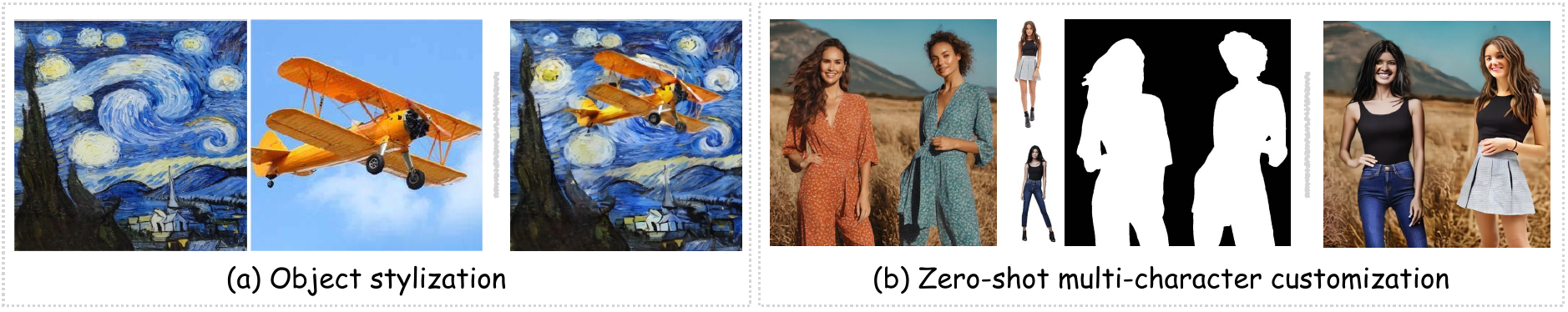}
\caption{\textbf{Other applications of \method,} including (a) object stylization and (b) zero-shot multi-character customization.}
\label{fig:applications}
\end{figure}

\subsubsection{Object stylization.}

During the image harmonization phase, the default prompts do not favor any particular style. However, if an object is composed onto a background that differs in style (for example, from a real plane to an oil-painting background as shown in Figure  \ref{fig:applications}), these prompts can be used to transfer the object to match the style of the background.

\subsubsection{Zero-shot multi-character customization.}

Animate Anyone\cite{hu2023animate} is a method that customizes images into videos by allowing zero-shot customization of a single character with a similar background. With the implementation of this method, it becomes possible to compose multiple customized characters together, thus enabling zero-shot multi-character customization.

\section{More Implementation Details}

\subsubsection{Optimization Steps.}
The best results in different cases are achieved through various optimization steps. Generally, we use 150 steps for object removal and 200 steps for image harmonization. However, for semantic image composition, the specific format of the conditions requires different numbers of steps. For instance, text requires 500 steps, while sketch and canny require 200 steps.

\subsubsection{Timestep Choice.}
According to our observations, different timesteps have varying levels of influence on the optimization results. During the object removal phase, we use timesteps ranging from 50 to 400 to enhance efficiency. For the image harmonization phase, timesteps between 50 and 950 are employed to achieve a more balanced outcome. In the semantic image composition phase, timesteps between 50 and 100 are used specifically for the final fifty optimization steps to ensure smoothness in the resulting image..

\subsubsection{T2I-Adapter Model.}
We utilize the T2I-Adapter, which was released by TencentARC\footnote{https://huggingface.co/TencentARC}, to apply the diffusion model to conditions in formats other than text. When it comes to image composition, sketch and canny are conditions more suitable than other formats, used for cases in our results.

\subsubsection{Running Times.}
The running times depend on the optimization steps chosen for a specific task. In general, when using an RTX 3090 with a float 16 precision, the first 50 steps take approximately 30 seconds, including preparation time for each phase. Subsequent sets of 50 steps take around 25 seconds.

\section{More Results}

\subsection{Object Removal}
\begin{figure}[htb]
    \centering
    \includegraphics[width=0.98\textwidth]{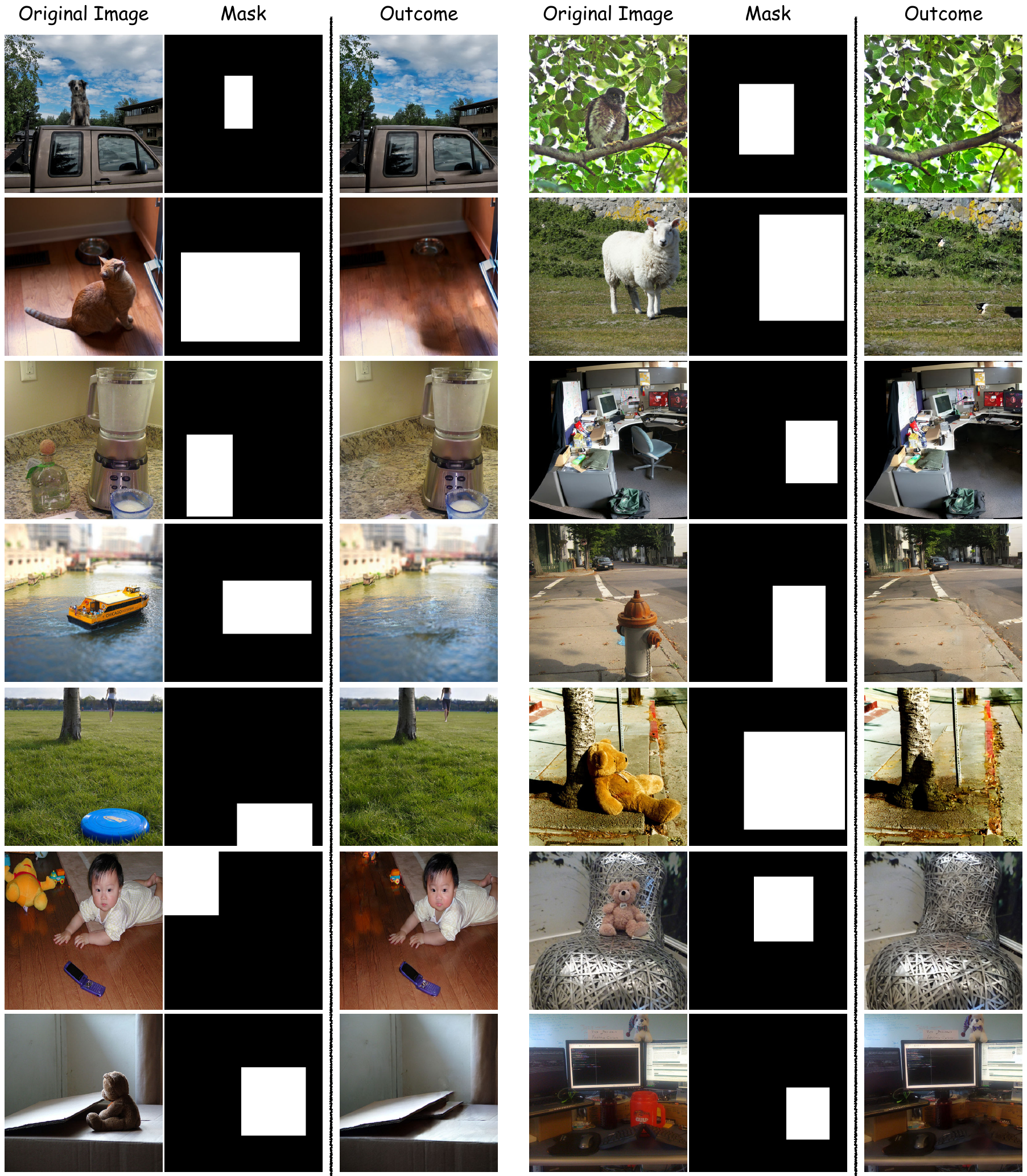}
    \caption{\textbf{More object removal results.} We show more object removal results in this figure. The first column is the original image, the second column is the mask of the object to be removed, the third column is the result of the object removal.}
    \label{fig:removal_supp}
\end{figure}
We show some more object removal results in Figure \ref{fig:removal_supp}. Our method can be widely applied to different types of objects and scenes, and can achieve good results in most cases.

\subsection{Image Harmonization}
\begin{figure}[htb]
    \centering
    \includegraphics[width=0.98\textwidth]{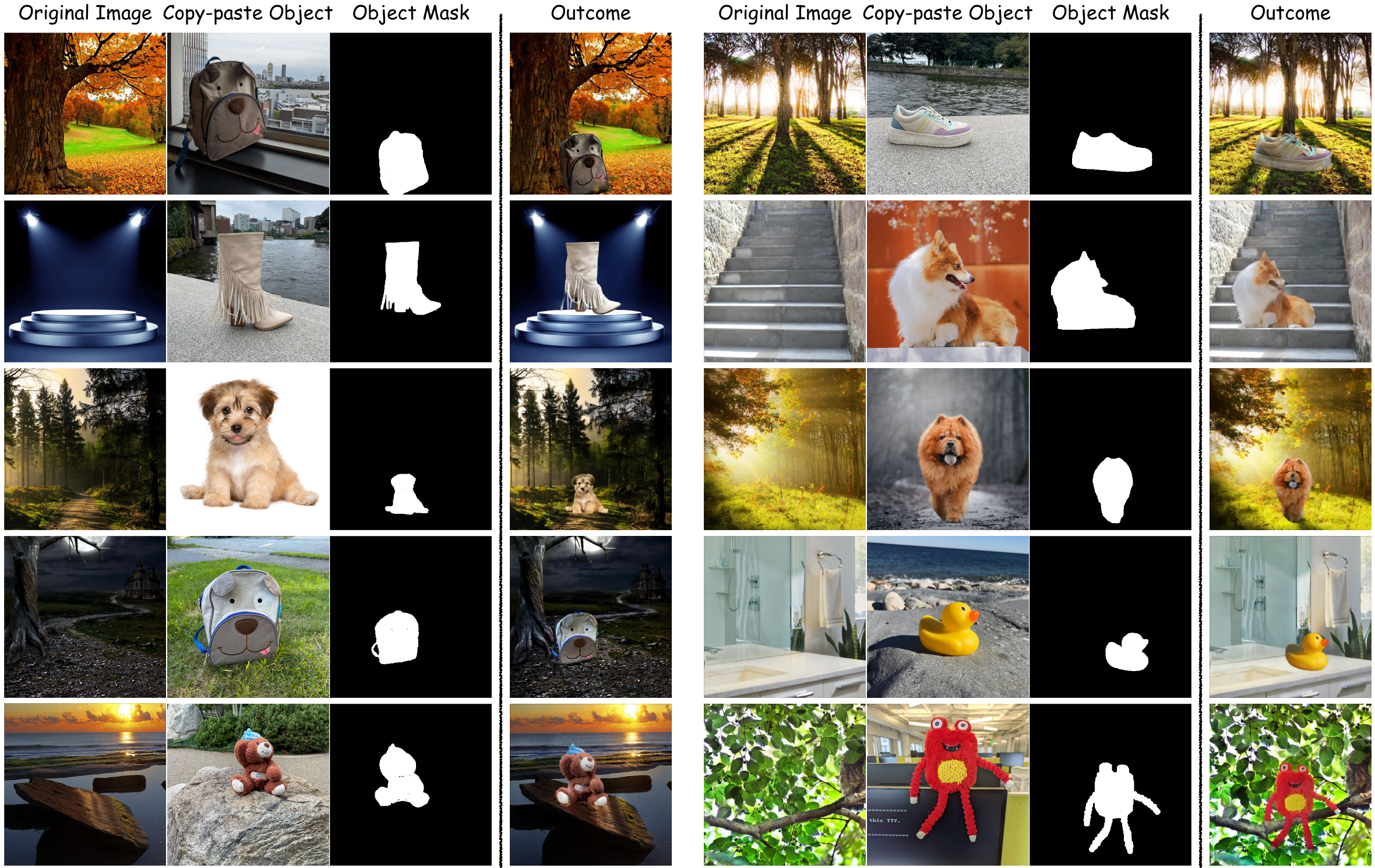}
    \caption{\textbf{More image harmonization results.} We show more image harmonization results in this figure. The first column is the original background image, the second column is the object to be pasted and harmonized, the third column is the mask of the object after being pasted, and the fourth column is the result of the image harmonization.}
    \label{fig:harmonization_supp}
\end{figure}
We show some more image harmonization results in Figure \ref{fig:harmonization_supp}. Our method automatically analyze the light and shadow of the environments and harmonize the object accordingly.

\subsection{Semantic Image Composition}
\begin{figure}[htb]
    \centering
    \includegraphics[width=0.98\textwidth]{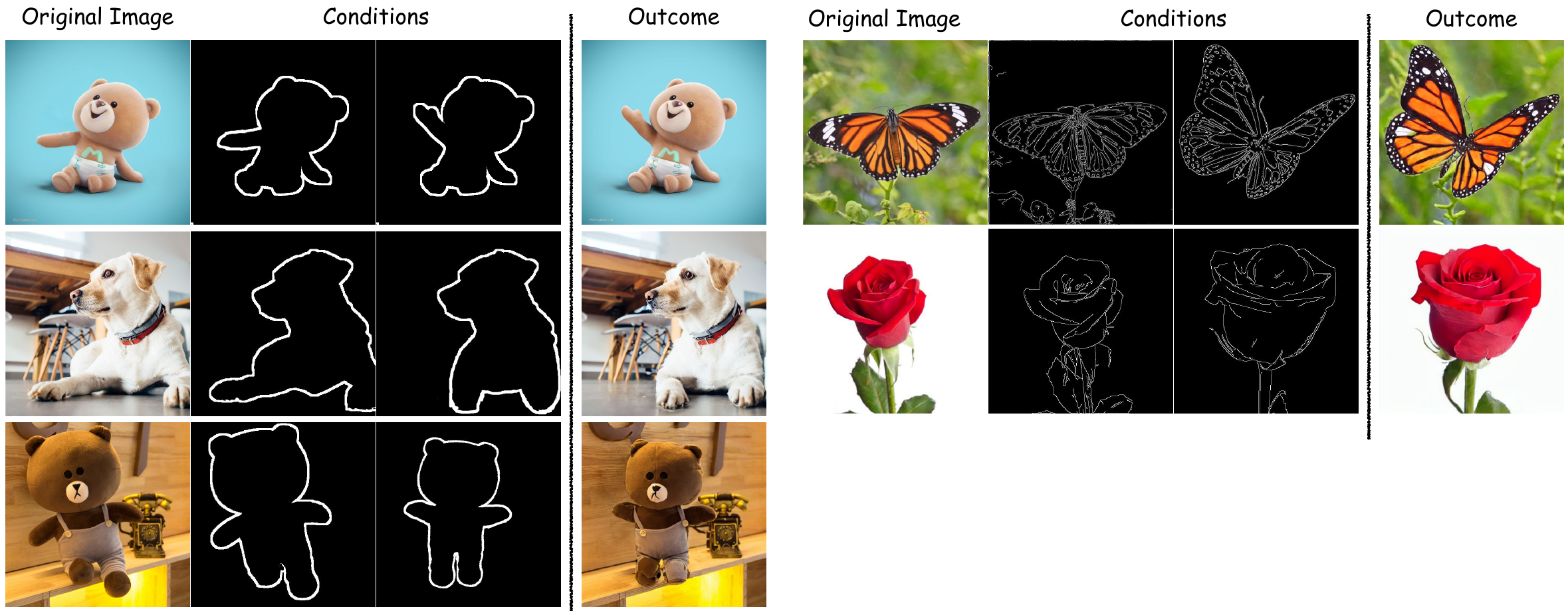}
    \caption{\textbf{More semantic image composition results.} We show more semantic image composition results in this figure. The left is the cases using sketches as conditions and the right is the cases using canny edges as conditions. For each side, the first column is the original image, the second column and the third column are the corresponded condition for original image and the condition for target image, and the fourth column is the result of the semantic image composition.}
    \label{fig:composition_supp}
\end{figure}
We show some more semantic image composition results in Figure \ref{fig:composition_supp}. Our method enables the use of various conditions as guidance to guide the composition process. In cases where more intricate texture or structure is desired, canny edges can be employed as conditions to achieve superior outcomes, as demonstrated in the right column.

\section{Plug-and-Play On other Diffusion Models}
\begin{figure}[htb]
    \centering
    \includegraphics[width=0.98\textwidth]{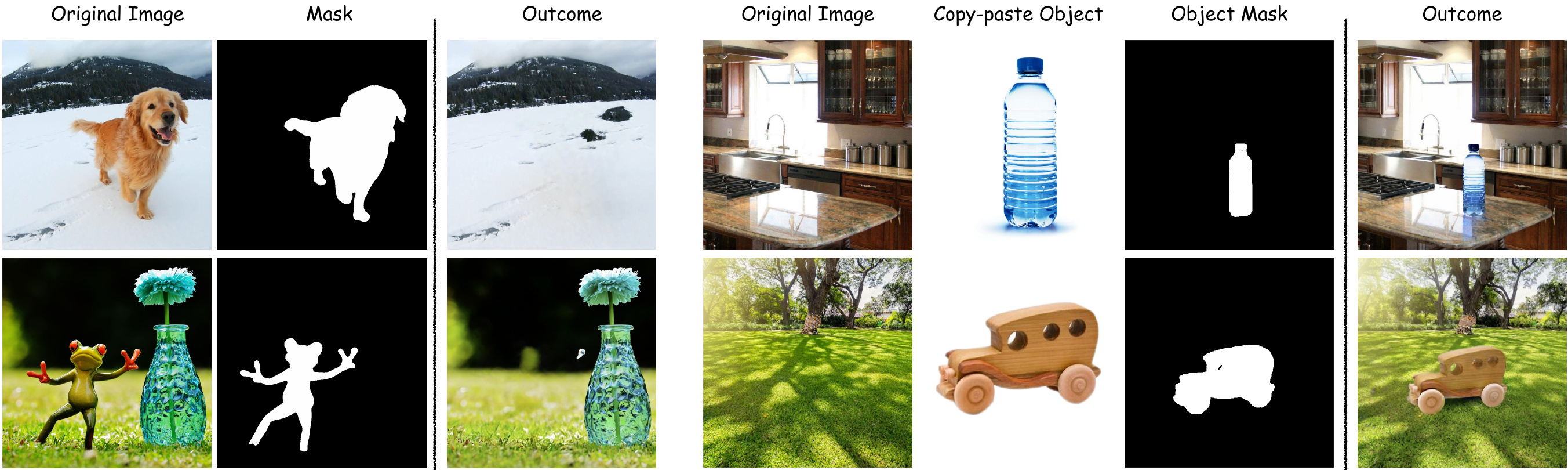}
    \caption{\textbf{\method on pre-trained SDXL.} We apply our method on pre-trained SDXL model. The left column is the results of object removal and the right column is the results of image harmonization.}
    \label{fig:sdxl}
\end{figure}
\subsubsection{Plug-and-Play on SDXL Model.} We apply \method to a pre-trained SDXL model\footnote{https://huggingface.co/stabilityai/stable-diffusion-xl-base-1.0}., and the results are displayed in Figure \ref{fig:sdxl}. Thanks to the exceptional prior of the SDXL model, the results are particularly impressive, especially in terms of image harmonization. As shown in the right column, it can be observed that the bottle's reflection on the table in the first case and the object's shadow in the second case are well integrated with the background through our method.

\section{Algorithm}

\subsection{Object Removal}
\begin{algorithm}[htb]
\caption{Object Removal}
\label{alg:removal}
\begin{algorithmic}[1]
\Require{Image $I$, Mask $M$, Diffusion model with parameter $\theta$, Optimization steps $S$. Get text embeddings $c_u, c_s, c_t$ from empty prompt and default prompts $P_s, P_t$. }{Given upper bound $t_{\rm max}$ and lower bound $t_{\rm min}$ for timesteps, guidance scale $w$, loss weight $\lambda_{\rm per}$ and learning rate $\eta$.}
\State $M' \gets {1 - M}$
\State $\mathbf{\hat{z}} \gets \text{Encode}(I)$
\State $\mathbf{z} \gets \mathbf{\hat{z}}$
\For{$s = 1, 2, \dots, S$}
    \State $t \gets \text{random}(t_{\rm min}, t_{\rm max})$
    \State $\alpha_t \gets \text{scheduler}(t)$
    \State $\epsilon \gets \mathcal{N}(0, \mathbf{I})$
    \State $\mathbf{z_t}, \mathbf{\hat{z_t}} \gets \sqrt{\alpha_t}\mathbf{z}+\sqrt{1-\alpha_t}\epsilon, \sqrt{\alpha_t}\mathbf{\hat{z}}+\sqrt{1-\alpha_t}\epsilon$ \Comment{random noise the latent}
    \State $\epsilon_{us}, \epsilon_{cs} \gets \epsilon_\theta(\mathbf{\hat{z}}, t, c_u), \epsilon_\theta(\mathbf{\hat{z}}, t, c_s)$
    \State $\epsilon_{ut}, \epsilon_{ct} \gets \epsilon_\theta(\mathbf{z}, t, c_u, M), \epsilon_\theta(\mathbf{z}, t, c_s, M)$ \Comment{mask guided calculation}
    \State $\epsilon_{s},\epsilon_{t} \gets \epsilon_{us} + w(\epsilon_{cs} - \epsilon_{us}),\epsilon_{ut} + w(\epsilon_{ct} - \epsilon_{ut})$
    \State $\mathcal{L} \gets {||\epsilon_s - \epsilon_t||}^2_2 + \lambda_{\rm per}\mathcal{L}_{\rm per}(I\otimes M', \text{Decode}(\mathbf{z})\otimes M')$
    \State $\mathbf{z} \gets \mathbf{z} - \eta\nabla_{\mathbf{z}}\mathcal{L}$
\EndFor \\
\Return The background image $\text{Decode}(\mathbf{z})$
\end{algorithmic}
\end{algorithm}
The pseudocode for our method in object removal phase is shown in Algorithm \ref{alg:removal}. The critical part is the calculation of the mask guided loss, which uses the mask for discarding semantic message during denoising of the target image.

\subsection{Image Harmonization}
\begin{algorithm}[htb]
\caption{Image Harmonization}
\label{alg:harmonization}
\begin{algorithmic}[1]
\Require{Image $I$, Mask $M$, Diffusion model with parameter $\theta$, Optimization steps $S$. Get text embeddings $c_u, c_s, c_t$ from empty prompt and default prompts $P_s, P_t$. }{Given upper bound $t_{\rm max}$ and lower bound $t_{\rm min}$ for timesteps, guidance scale $w$, loss weight $\lambda_{\rm bak},\Lambda_{\rm for}$ and learning rate $\eta$.}
\State $M' \gets {1 - M}$
\State $\mathbf{\hat{z}} \gets \text{Encode}(I)$
\State $\mathbf{z} \gets \mathbf{\hat{z}}$
\For{$s = 1, 2, \dots, S$}
    \State $t \gets \text{random}(t_{\rm min}, t_{\rm max})$
    \State $\alpha_t \gets \text{scheduler}(t)$
    \State $\epsilon \gets \mathcal{N}(0, \mathbf{I})$
    \State $\mathbf{z_t}, \mathbf{\hat{z_t}} \gets \sqrt{\alpha_t}\mathbf{z}+\sqrt{1-\alpha_t}\epsilon, \sqrt{\alpha_t}\mathbf{\hat{z}}+\sqrt{1-\alpha_t}\epsilon$ \Comment{random noise the latent}
    \State $\epsilon_{us}, \epsilon_{cs} \gets \epsilon_\theta(\mathbf{\hat{z}}, t, c_u), \epsilon_\theta(\mathbf{\hat{z}}, t, c_s)$
    \State $\epsilon_{ut}, \epsilon_{ct} \gets \epsilon_\theta(\mathbf{z}, t, c_u), \epsilon_\theta(\mathbf{z}, t, c_s)$
    \State $\epsilon_{s},\epsilon_{t} \gets \epsilon_{us} + w(\epsilon_{cs} - \epsilon_{us}),\epsilon_{ut} + w(\epsilon_{ct} - \epsilon_{ut})$
    \State $\mathcal{L}_{\rm bak} \gets \mathcal{L}_{\rm per}(I\otimes M', \text{Decode}(\mathbf{z})\otimes M')$
    \State $\mathcal{L}_{\rm for} \gets \mathcal{L}_{\rm per}(I\otimes M, \text{Decode}(\mathbf{z})\otimes M)$
    \State $\mathcal{L} \gets {||\epsilon_s - \epsilon_t||}^2_2 + \lambda_{\rm bak}\mathcal{L}_{\rm bak} + \lambda_{\rm for}\mathcal{L}_{\rm for}$ \Comment{compose all loss}
    \State $\mathbf{z} \gets \mathbf{z} - \eta\nabla_{\mathbf{z}}\mathcal{L}$
\EndFor \\
\Return The harmonized image $\text{Decode}(\mathbf{z})$
\end{algorithmic}
\end{algorithm}
The pseudocode for our method in image harmonization is presented in Algorithm \ref{alg:harmonization}. This section balances various losses to find a tradeoff between object identity, background features, and overall harmony.

\subsection{Semantic Image Composition}
\begin{algorithm}[htb]
\caption{Semantic Image Composition}
\label{alg:composition}
\begin{algorithmic}[1]
\Require{Image $I$, Mask $M$, Diffusion model with parameter $\theta$, Optimization steps $S$ and $\tau_s, \tau_l$ for restriction of step and layer to begin replacing. Get text embeddings $c_u, c_s, c_t$ from empty prompt and prompts $P_s, P_t$, and features $f_s, f_t$ from conditions $C_s, C_t$ through pre-trained T2I-Adapters. }{Given upper bound $t_{\rm max}$ and lower bound $t_{\rm min}$ for timesteps, guidance scale $w$ and learning rate $\eta$.}
\State $M' \gets {1 - M}$
\State $\mathbf{\hat{z}} \gets \text{Encode}(I)$
\State $\mathbf{z} \gets \mathbf{\hat{z}}$
\For{$s = 1, 2, \dots, S$}
    \State $t \gets \text{random}(t_{\rm min}, t_{\rm max})$
    \State $\alpha_t \gets \text{scheduler}(t)$
    \State $\epsilon \gets \mathcal{N}(0, \mathbf{I})$
    \State $\mathbf{z_t}, \mathbf{\hat{z_t}} \gets \sqrt{\alpha_t}\mathbf{z}+\sqrt{1-\alpha_t}\epsilon, \sqrt{\alpha_t}\mathbf{\hat{z}}+\sqrt{1-\alpha_t}\epsilon$ \Comment{random noise the latent}

    \State $\epsilon_{us}, \{Q_{us}, K_{us}, V_{us}\} \gets \epsilon_\theta(\mathbf{\hat{z}}, t, c_u; f_s)$ \Comment{use condition features}
    \State $\epsilon_{cs}, \{Q_{cs}, K_{cs}, V_{cs}\} \gets \epsilon_\theta(\mathbf{\hat{z}}, t, c_s; f_s)$
    \If{$s > \tau_s$}\\
        \Comment{use condition features and replace self-attention features of layer index $l > \tau_l$}
        \State $\epsilon_{ut} \gets \epsilon_\theta(\mathbf{z}, t, c_u; f_t, \{Q_{us}, K_{us}, V_{us}\})$ 
        \State $\epsilon_{ct} \gets \epsilon_\theta(\mathbf{z}, t, c_s; f_t, \{Q_{cs}, K_{cs}, V_{cs}\})$
    \Else
        \State $\epsilon_{ut}, \epsilon_{ct} \gets \epsilon_\theta(\mathbf{z}, t, c_u; f_t), \epsilon_\theta(\mathbf{z}, t, c_s; f_t)$
    \EndIf

    \State $\epsilon_{s},\epsilon_{t} \gets \epsilon_{us} + w(\epsilon_{cs} - \epsilon_{us}),\epsilon_{ut} + w(\epsilon_{ct} - \epsilon_{ut})$
    \State $\mathcal{L} \gets {||\epsilon_s - \epsilon_t||}^2_2$
    \State $\mathbf{z} \gets \mathbf{z} - \eta\nabla_{\mathbf{z}}\mathcal{L}$
\EndFor \\
\Return The composed image $\text{Decode}(\mathbf{z})$
\end{algorithmic}
\end{algorithm}
The pseudocode for our method in semantic image composition is demonstrated in Algorithm \ref{alg:composition}. The key aspect is the utilization of condition features to guide the transformation and the replacement of the self-attention features of the target image, which forms the core of the semantic image composition phase.

\section{Discussion}

\subsection{Limitations}
The first limitation concerns the object removal phase. Through the use of mask guided loss, the pipeline replaces the semantic information of the object with that of the background. However, if the mask is too large, the remaining background information may not be enough to accurately reconstruct the entire background, leading to the creation of artifacts. Additionally, it is important that the mask fully covers the object to be removed; otherwise, certain portions of the object may still be visible in the final result. In situations where there are similar objects present in the background, the pipeline may mistakenly replace the removed object with these similar objects, as they share a similar semantic message.

The second limitation pertains to the image harmonization phase. Although the pipeline achieves excellent results in terms of light and shadow, it struggles to strike a balance between the object's features and the overall naturalness when there is a significant contrast between the object and the background. For instance, when dealing with an object that has dark shadows against a bright background.

The third limitation relates to the semantic image composition phase. The quality of the output is partially influenced by the format and quality of the input conditions. When it comes to text prompts, the pipeline can only generate subtle variations. As for sketches, certain details are challenging to render realistically. Canny edges appear to be the most suitable format for conditions, but they are less accessible and more intricate.

\subsection{Future Work}
\method enables flexible composition among different objects and backgrounds by utilizing pre-trained diffusion models, without the need for additional training. In the future, we plan to expand our method to cover more composition tasks and further explore the potential of the pipeline. We also intend to investigate the feasibility of applying our method to video models and other generative models. Additionally, we will improve the user-friendliness and efficiency of the pipeline in future updates.

\subsection{Negative Impact}
Our \method aims to utilize the prior knowledge of pre-trained diffusion models and extend their use to tasks beyond their original purpose. However, it is important to acknowledge the potential for malicious applications of our method, such as generating deceptive images that composing real individuals with fabricated surroundings for the purpose of misinformation and disinformation. This is a common issue with generative models.

One possible way to address the negative impact is to adopt methods similar to that  proposed by Pham et al. \cite{pham2024circumventing}. These methods leverage the capability of diffusion models to identify fake images and help prevent the abuse of our method. Furthermore, it is crucial to be mindful of employing unseen watermarks and other techniques to authenticate images in order to prevent the misuse of our method.